\documentclass[acmsmall]{acmart}

\usepackage{ifpdf}

\usepackage{hyperref}
\usepackage{amsmath}
\DeclareMathOperator*{\argmax}{argmax}
\DeclareMathOperator*{\argmin}{argmin}
\usepackage{algorithm}
\usepackage{algorithmicx}
\usepackage{algpseudocode}
\usepackage{tabularx}
\usepackage{array}
\usepackage{url}
\usepackage{makecell}
\usepackage{booktabs}
\usepackage{multirow}
\usepackage{caption}
\AtBeginDocument{%
  \providecommand\BibTeX{{%
    \normalfont B\kern-0.5em{\scshape i\kern-0.25em b}\kern-0.8em\TeX}}}

\copyrightyear{Published in The \textit{IEEE Transactions on Neural Networks and Learning Systems}}

\graphicspath{{./img/}}
\begin{document}

\title{Imitation Learning: Progress, Taxonomies and Challenges}

\author{Boyuan Zheng}

\email{14055661@student.uts.edu.au}
\orcid{0000-0003-1223-9230}

\author{Sunny Verma}
\author{Jianlong Zhou}
\author{Ivor Tsang}
\author{Fang Chen}
\affiliation{%
  \institution{University of Technology Sydney}
  \streetaddress{PO Box 123}
  \city{Sydney}
  \state{New South Wales}
  \country{Autralia}
  \postcode{2007}
}

\renewcommand{\shortauthors}{Boyuan et. al.}

\begin{abstract}
Imitation learning aims to extract knowledge from human experts' demonstrations or artificially created agents in order to replicate their behaviours. Its success has been demonstrated in areas such as video games, autonomous driving, robotic simulations and object manipulation. However, this replicating process could be problematic, such as the performance is highly dependent on the demonstration quality, and most trained agents are limited to perform well in task-specific environments. In this survey, we provide a systematic review on imitation learning. We first introduce the background knowledge from development history and preliminaries, followed by presenting different taxonomies within Imitation Learning and key milestones of the field. We then detail challenges in learning strategies and present research opportunities with learning policy from suboptimal demonstration, voice instructions and other associated optimization schemes.
\end{abstract}




\maketitle

\section{Introduction}
%
%
%

Imitation learning (IL), also known as learning from demonstration, makes responses by mimicking behavior in a relatively simple approach. It extracts useful knowledge to reproduce the behavior in the environment which is similar to the demonstrations'. 
The presence of IL facilitates the research on autonomous control system and designing artificially intelligent agents, as it demonstrates good promise in real-world scenario and efficiency to train a policy. 
Recent developments in machine learning field like deep learning, online learning and Generative Adversarial Network (GAN)\hspace{1pt}\cite{goodfellow2014generative} make further improvement on IL, not only alleviating existing problems like dynamic environment, frequent inquiries and high-dimensional computation, but also achieving faster convergence, more robust to the noise and more sample-efficient learning process. These improvements of IL promote the applications in both continuous and discrete control domains. 
For example, in the continuous control domain, imitation learning could be applied to autonomous vehicle manipulation to reproduce appropriate driving behavior in a dynamic environment\cite{pomerleauALVINNAutonomousLand1989,pomerleauEfficientTrainingArtificial1991,georgeImitationLearningEnd2018,kebriaDeepImitationLearning2020,zhouModelingCarFollowingBehaviors2020,codevillaEndtoEndDrivingConditional2018,chenLearningCheating2019,buhlerDrivingGhostsBehavioral2020}. In addition, imitation learning is also applied to robotic, ranging from basic grabbing and placing to surgical assistance\cite{finnGuidedCostLearning2016,lioutikovLearningMovementPrimitive2017,osaOnlineTrajectoryPlanning2014,osaOnlineTrajectoryPlanning2018,nairCombiningSelfSupervisedLearning2017,sunDeeplyAggreVaTeDDifferentiable2017,zhangDeepImitationLearning2018,osaGuidingTrajectoryOptimization2017}. In the discrete control domain, imitation learning makes contribution to fields like game theory\cite{rossEfficientReductionsImitation2010,hesterDeepQlearningDemonstrations2017,aytarPlayingHardExploration2018,edwardsImitatingLatentPolicies2019}, navigation tasks\cite{husseinDeepImitationLearning2018,wangRobustImitationDiverse2017,shouOptimalPassengerseekingPolicies2020}, cache management\cite{liu2020imitationcache} and so on. 

It is worth noting that the demonstrations could be gathered either from human experts or artificial agents. In most cases, the demonstration is collected from human experts, but there are also some studies that obtain the demonstration through another artificial agent. For example, Chen et al.\cite{chenLearningCheating2019} proposed a teacher-student training structure, they train a teacher agent with additional information and use this trained agent to teach a student agent without additional information. This process is not redundant, using the demonstration from other agent benefits the training process as student agents can rollout their own policy by frequently querying trained agents and learn policies from similar configurations while classic IL needs to overcome the kinematic shifting problem.

IL has a close relationship with Reinforcement Learning (RL). Both IL and RL commonly solve the problem under Markov Decision Process, and improvements like TRPO\cite{schulmanTrustRegionPolicy2017} in RL could benefit IL as well, but they reproduce the behavior in a different manner. In comparing to RL, IL is more \textit{efficient}, \textit{accessible}, and \textit{human-interactive}. In terms of \textit{efficiency}, comparing with trial and error, the IL agents usually spend less time to produce the desired behavior by using the demonstrations as guidance. In terms of \textit{accessibility}, achieving autonomous behavior in the RL approach requires human experts who are familiar with the problem setting, together with hard-coded reward functions which could be impractical and non-intuitive in some settings. For example, people learn to swim and walk almost from demonstration instead of math functions, and it is hard to formulate these behavior mathematically. IL also prompts interdisciplinary integration, experts who are novice to programming can contribute to the design and evaluating paradigms. In terms of \textit{human-interaction}, IL highlights human's influence through providing demonstration or preference to accelerate the learning process, which efficiently leverages and transfers the experts' knowledge. Although IL presents the above merits, it also faces challenges and opportunities, and this content will be detailed in the following sections.

This survey is organized as follows:
\begin{itemize}
\item {\bf{Systematic review} \textnormal{This survey presents research in imitation learning under categories \textit{behavioural cloning vs. inverse reinforcement learning} and \textit{model-free vs. model-based}. It then summarizes IL research into two new categories namely \textit{low-level tasks vs. high-level tasks} and \textit{BC vs. IRL vs. Adversarial Structured IL}, which are more adapted to the development of IL.}}
\item \bf{Background knowledge} \textnormal{A comprehensive description of IL's evolution is presented in Section \ref{Background}, followed by fundamental knowledge in Section \ref{pre} and the most common learning framework in Sections \ref{main topic}.}
\item \bf{Future direction} \textnormal{This survey presents the remaining challenges of IL, like learning diverse behavior, leveraging various demonstration and better representation. Then we discuss the future directions with respect to methods like transfer learning and importance sampling.}
\end{itemize}

\section{Background}\label{Background}

\begin{figure*}[t]
      \centering
      \includegraphics[width=0.9\textwidth]{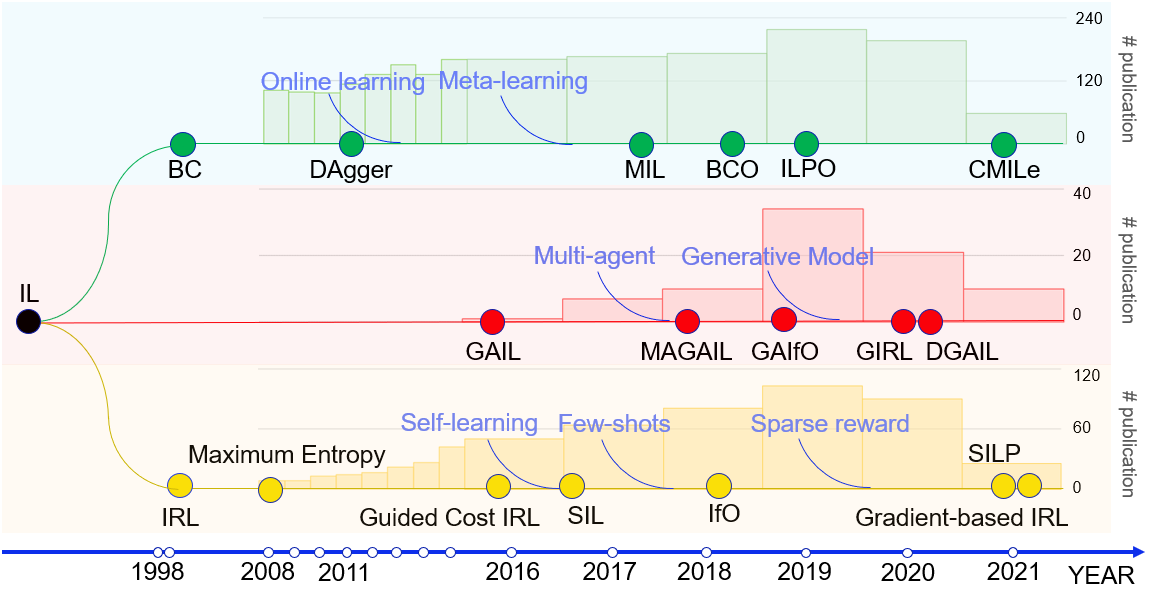}
      \caption{Featured approaches and annual publication numbers for each class of approaches. The blue text indicates some of the most active research topics in IL and the background histogram plot is the number of annual publications. The data was collected from Web of Science until 31 May 2021, filtered by setting up each class and their abbreviation as keywords (like ``Behavioural Cloning OR BC", only cover records within computer science). }
      \label{fig_keyword}
\end{figure*}

One of the earliest well-known research on IL is the Autonomous Land Vehicle In a Neural Network (ALVINN) project at Carnegie Mellon University proposed by Pomerleau\cite{pomerleauALVINNAutonomousLand1989}. In 1998, a formal definition of Inverse Reinforcement Learning (IRL) was proposed by Russell\cite{russelllearning1998}. Inverse reinforcement learning aims to recover reward function from demonstrations. A year after, a formal definition of another important category -- Behavioural Cloning (BC) was proposed in\cite{bainFrameworkBehaviouralCloning1999}. 
BC works in a supervised learning fashion and seeks to learn a policy that builds a direct mapping between states and actions, then output a control strategy for control tasks. Although BC demonstrates significant advantage in efficiency, it also suffers from various problems. In 2010, SMiLe\cite{rossEfficientReductionsImitation2010} was proposed, it mixed a new policy $\hat{\pi}^{n+1}$ with a fixed probability $\alpha$ as next policy, this method promotes the development of IL and set up the foundation for the later proposed DAgger\cite{rossReductionImitationLearning2011}. 
DAgger was proposed by Ross et al. and it updates the dataset in each iteration and trains a new policy in the subsequent iteration based on the updated dataset. Compared with previous methods like SMILe \cite{rossEfficientReductionsImitation2010} and SEARN \cite{daumeiiiSearchbasedStructuredPrediction2009a}, DAgger alleviates the problem on the unseen scenario and achieve data-efficiency. Later research like\cite{rossReinforcementImitationLearning2014,sunDeeplyAggreVaTeDDifferentiable2017,liu2020imitationcache} were proposed to make improvements on DAgger. Besides DAgger and its derivatives, other BC methods also make contribution to the development of IL like MMD-IL\cite{kimMaximumMeanDiscrepancy2013}, LOLS\cite{changLearningSearchBetter2015}.
As for applications, one of the notable applications of BC was proposed by Abbeel et al.\cite{abbeelAutonomousHelicopterAerobatics2010}, a model-free BC method on autonomous helicopter project, developed an open-loop iterative learning control. Another famous BC application was an autonomous surgical knot-tying robotic proposed by Osa et al.\cite{osaOnlineTrajectoryPlanning2018}, which achieved online trajectory planning and updating in a dynamic system. Besides these real-world applications, BC was also implemented into other research fields like cybernetics, for example, DAgger was used for scheduling in \cite{wang2021minimizing} and Liu et al. leveraged Belagy's optimal policy (proof-of-concept) as oracle to solve the cache replacement problem by predicting reuse distance when cache miss happens\cite{liu2020imitationcache}. 

In terms of IRL, Ziebart et al.\cite{ziebart2008maximum} proposed Maximum Entropy IRL, which uses maximum entropy distribution to develop a convex procedure for good promise and efficient optimization. This method played a pivotal role in the development of subsequent IRL and GAIL. In 2016, Finn et al.\cite{finnGuidedCostLearning2016} made significant contributions to IRL and proposed a model-based IRL method called guided cost learning, neural network is used for representing cost to enhance expressive power, combining with sample-based IRL to handle the unknown dynamics. Later in 2017, Hester et al. proposed DQfD\cite{hesterDeepQlearningDemonstrations2017} which uses small amount of demonstration to significantly accelerate the training process by doing pre-training to kick-off and learning from both demonstration and self-generated data. Later methods like T-REX\cite{brownExtrapolatingSuboptimalDemonstrations2019}, SQIL\cite{reddySQILImitationLearning2019}, SILP\cite{luo2021self} make improvements on IRL from different aspects.

Another novel method called Generative Adversarial Imitation Learning (GAIL), it was proposed in 2016 by Ho and Ermon\cite{hoGenerativeAdversarialImitation2016} and became one of the hot topics in IL. Later research like\cite{wangRobustImitationDiverse2017,stadieThirdPersonImitationLearning2017,dingGoalconditionedImitationLearning2019,kinoseIntegrationImitationLearning2020} were proposed inspired by GAIL and other generative models were gradually adopted in IL. Besides GAIL, another important research direction is inspired by Stadie et al.\cite{stadieThirdPersonImitationLearning2017}. Since first-person demonstrations are hard to obtain in practice, and people usually learn by observing the demonstration of others through the perspective of a third party, learning from third-person viewpoint demonstrations was proposed. The change of viewpoint facilitates the following research like\cite{edwardsImitatingLatentPolicies2019,brownExtrapolatingSuboptimalDemonstrations2019}, which includes IfO\cite{liuImitationObservationLearning2018}. IfO focus on simplifying input to use raw video only (i.e. no longer use state-action pairs), many following methods advocate this new setting. These methods measure the distance between observations to replace the need for ground-truth actions and widen the available input for training, foe example, using YouTube videos for training\cite{aytarPlayingHardExploration2018}. Other interesting research fields like meta-learning\cite{finnModelAgnosticMetaLearningFast2017,duanOneShotImitationLearning2017,huTwoStageModelAgnosticMetaLearning2020}, multi-agent learning\cite{zhanGeneratingMultiAgentTrajectories2019} are also thrived because of the development of IL. Figure \ref{fig_keyword} shows some featured approaches and annual publication numbers for each class and focuses on the research after 2016, it shows that the class of BC(Behavioural Cloning) has maintained a stable increment in publications, while the research in the class of Adversarial Structured IL and IRL(Inverse Reinforcement Learning) have grown rapidly due to the recent advance in other research fields like deep learning.

\section{Preliminary Knowledge}\label{pre}
This section provides some basic concepts for better understanding of the IL methodology.

In IL, the demonstrated trajectories are commonly represented as pairs of states $s$ and actions $a$, sometimes other parameters such as high-level commands and conditional goals will also be included to form the dataset. The way to collect the dataset could be either online or offline. Offline IL prepares the dataset in advance and obtains policies from the dataset while involves fewer interactions with the environment. This could be beneficial when interacting with the environment is expensive or risky. Contrary to offline learning, online learning assumes the data would be accessible in sequence and uses this updated data to learn the best predictor for future data. This method facilitates imitation learning to be more robust in a dynamic system. For example, in\cite{rossReductionImitationLearning2011,osaOnlineTrajectoryPlanning2014,osaOnlineTrajectoryPlanning2018}, online learning is used in surgical robotics. The online learning agent will provide a policy in iteration \textit{n}, then the opponent will choose a loss function \textit{$l_n$} based on current policy and the new observed loss will affect the choice of next iteration \textit{$n+1$}'s policy. The performance is measured through regret, i.e.
\[
\sum_{n=1}^{N}l_n(\pi_n)-\min_{\pi\in\Pi}\sum_{n=1}^{N}l_n(\pi),
\]
and the loss function could vary from iteration to iteration. One of the most common ways to calculate loss is Kullback-Leibler (KL) Divergence. KL Divergence measures the difference between 2 probability distribution, i.e.,
\[
D_{KL}(p(x)\parallel q(x)) = \int p(x)\text{ln}\frac{p(x)}{q(x)}dx.
\]
KL divergence is not symmetric, i.e., $D_{KL}(p(x)\parallel q(x)) \neq D_{KL}(q(x)\parallel p(x))$. Many algorithms such as\cite{schulmanTrustRegionPolicy2017,bhattacharyyaMultiAgentImitationLearning2018} use KL divergence as the loss function as it could be useful when dealing with the stochastic policy learning problem.

For many methods, especially those under the class of IRL and Adversarial structured IL, the environment is modeled as Markov Decision Process(MDP). MDP is the process satisfying the property that the next state $s_{t+1}$ only depends on the current state $s_t$ at any time $t$. Typically, a MDP is defined as a tuple ($\mathcal{S}$,$\mathcal{A}$,$\mathcal{P}$,$\gamma$,$\mathcal{D}$,$\mathcal{R}$), where $\mathcal{S}$ is the finite set of states, $\mathcal{A}$ is the corresponding set of actions, $\mathcal{P}$ is the set of state transition probabilities and the successor states $s_{t+1}$ is drawn from this transition model, i.e. $s_{t+1}=P(\cdot|s_t,a_t)$, $\gamma \in [1,0)$ is the discount factor, $\mathcal{D}$ is the set of initial state distribution and $\mathcal{R}$ is the reward function $\mathcal{S} \mapsto \mathbb{R}$, and in IL setting, the reward function is not available. The Markov property assists imitation learning to simplify the input since the earlier state is helpless to determine the next state. The use of MDP inspires research to make use of other MDP variants to solve various problems, for example, Partially Observable MDP is used to model the scheduling problem in \cite{wang2021minimizing} and Markov games is used in multi-agent scenario\cite{songMultiAgentGenerativeAdversarial2018}. 

The learning process of IL could be either on-policy or off-policy (there exists research using a hierarchical combination of these two\cite{chenLearningCheating2019}). On-policy learning estimates the return and updates the action using the same policy, the agent adopting on-policy will pick actions by themselves and rollout their own policy while training; Off-policy learning estimates the return and chooses the action using different policy, the agent adopting off-policy will update their policy greedily and imitate action with the help of other sources. Some recent IL research such as \cite{Zuo2020offpolicyRAIL,sasaki2018sample,blonde2019sample} advocates off-policy actor-critic architecture to optimize the agent policy and achieve sample efficiency comparing with on-policy learning.

\section{Categorization and Frameworks} \label{cat_fra}

\begin{figure*}[t]
      \centering
      \includegraphics[width=0.9\textwidth]{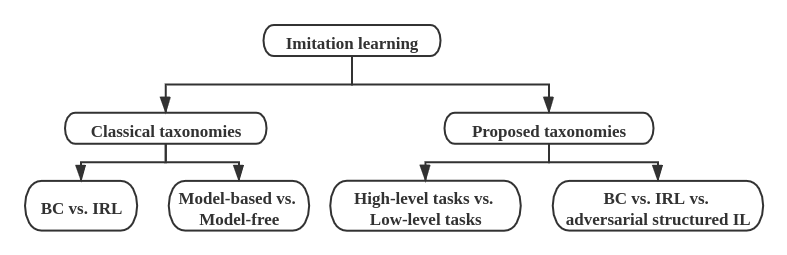}
      \caption{Taxonomies in this review.}
      \label{fig_tax}
\end{figure*}

In this section, four kinds of taxonomies are presented (see Figure \ref{fig_tax}). The first two taxonomies (BC vs. IRL and model-free vs. model-based) follow the classifications in\cite{osaAlgorithmicPerspectiveImitation2018,torabirecent2019} and the other two (Low=level Manipulation Tasks vs. High-Level Tasks and BC vs. IRL vs. adversarial structured IL are new proposed taxonomies. 

\subsection{Behavioural Cloning vs. Inverse Reinforcement Learning} 

\begin{table}[t]
\caption{Categorization of IL: BC vs. IRL}
\label{table_bc_irl}
\centering
\begin{tabular}{@{}cl@{}}
\toprule
Classes & Examples and Publications\\ \midrule
\multirow{4}{*}{Behavioural Cloning} & Few-shots learning\cite{duanOneShotImitationLearning2017} \\ \cmidrule(l){2-2} 
 & Input optimization\cite{chenLearningCheating2019} \\ \cmidrule(l){2-2} 
 & Latent policy learning\cite{lynchLearningLatentPlans2019} \\ \cmidrule(l){2-2} 
 & Real-world application\cite{zhangDeepImitationLearning2018} \\ \midrule
\multirow{4}{*}{Inverse Reinforcement Learning} & Improving efficiency\cite{brownExtrapolatingSuboptimalDemonstrations2019} \\ \cmidrule(l){2-2} 
 & Raw video as inputs\cite{sermanetTimeContrastiveNetworksSelfSupervised2018} \\ \cmidrule(l){2-2} 
 & Adversarial structured\cite{sunProvablyEfficientImitation2019} \\ \cmidrule(l){2-2} 
 & Sparse reward problem\cite{nairOvercomingExplorationReinforcement2018} \\ \bottomrule
\end{tabular}
\end{table}

IL is conventionally divided into BC and IRL. These two classes flourish by combining various techniques and then extend into different domains. Generally speaking, BC and IRL methods use different methodology to reproduce the expert behavior. BC commonly uses a direct mapping from the states to the actions, while IRL tries to recover the reward function from the demonstrations. This difference could be why BC methods are commonly applied to real-world problems while most IRL methods still do simulations in the environment with less invention.

Compared with direct mapping, recovering a reward function needs stronger computational power and technologies to obtain the unique reward function and solve the sparse reward problem. The inner loop reinforcement learning could also cause IRL methods to be impractical in real-world problems. For the computational problem, recent development in GPU gradually alleviate the problem of high-dimensional computation; for the technology aspect, recent algorithms like Trust Region Policy Optimization\cite{schulmanTrustRegionPolicy2017} and attention models\cite{hufinegrained2020} provide more robust and efficient approaches for IRL methods; as for the sparse reward function, Hindsight Experience Replay\cite{andrychowicz2017hindsight} is commonly adopted for this problem. On the other hand, BC also suffers from the ``compounding error"\cite{rossReductionImitationLearning2011} where a small error could destroy the final performance. Besides these problems, other problems like better representation and diverse behavior learning are still open, many approaches are proposed for these problems, such as\cite{wangRobustImitationDiverse2017,liu2019ahng,hussein2021robust}. 

Table \ref{table_bc_irl} lists some of the recent research in IL categorized into BC and IRL. Recent BC methods mainly focus on the topics such as: meta-learning that the agent is learning to learn by pretraining on a broader range of behaviors\cite{duanOneShotImitationLearning2017}; combining BC with other technique like VR equipment\cite{zhangDeepImitationLearning2018}. On the other hand, recent IRL methods mainly focus on the topics such as: extending GAIL with other methods or problem settings\cite{dingGoalconditionedImitationLearning2019}; recovering reward function from raw videos\cite{aytarPlayingHardExploration2018}; developing more efficient model-based IRL approaches by using the current development in reinforcement learning like TRPO\cite{schulmanTrustRegionPolicy2017} and HER\cite{andrychowicz2017hindsight}.

\subsection{Model-Based vs. Model-Free} 

\begin{table}[t]
\caption{Categorization of IL: Model-based vs. Model-free}
\label{table_mb_mf}
\centering
\begin{tabular}{@{}c@{\hskip 0.3in}l@{}}
\toprule
Classes & Examples and Publications\\ \midrule
\multirow{2}{*}{Model-based IL} & Forward model\cite{edwardsImitatingLatentPolicies2019,finnGuidedCostLearning2016} \\ \cmidrule(l){2-2} 
 & Inverse model\cite{nairCombiningSelfSupervisedLearning2017} \\ \midrule
\multirow{3}{*}{Model-free IL} & BC method\cite{lynchLearningLatentPlans2019} \\ \cmidrule(l){2-2} 
 & Reward engineering\cite{brownExtrapolatingSuboptimalDemonstrations2019} \\ \cmidrule(l){2-2} 
 & Adversarial style\cite{torabiGenerativeAdversarialImitation2019} \\ \bottomrule
\end{tabular}
\end{table}

Another classical taxonomy divides IL into model-based and model-free methods. The main difference between these two classes is whether the algorithm adopts a forward model to learn from the environmental context/dynamics. Before GAIL\cite{hoGenerativeAdversarialImitation2016} was proposed, most IRL methods are developed in the model-based setting because IRL methods commonly involve iterative algorithms evaluate the environment, while BC methods are commonly model-free since the low-level controller is commonly available. After GAIL was proposed, various adversarial structured IL are proposed following the GAIL's model-free setting.
Although learning from the environment sounds beneficial for all kinds of methods, it might not be necessary for a given problem setting or impractical to apply. Integrating environment context/dynamics could obtain more useful information so that the algorithm can achieve data-efficiency and feasibility, while the drawback is learning the model is expensive and challenging. For example, in robotics, the equipment is commonly precise, the spatial position, velocity and other parameters could be easily obtained, the system dynamics might provides relatively little help to reproduce the behavior. On the other hand, in autonomous car tasks, the system dynamics might be crucial to avoid hitting pedestrians. In this case, the choice of model-free or model-based depends on the tasks. Table \ref{table_mb_mf} lists some of the recent research topics in IL categorized into model-based and model-free.

\subsection{Low-Level Tasks vs. High-Level Tasks} 
\begin{figure}[t]
      \centering
      \includegraphics[width=0.9\textwidth]{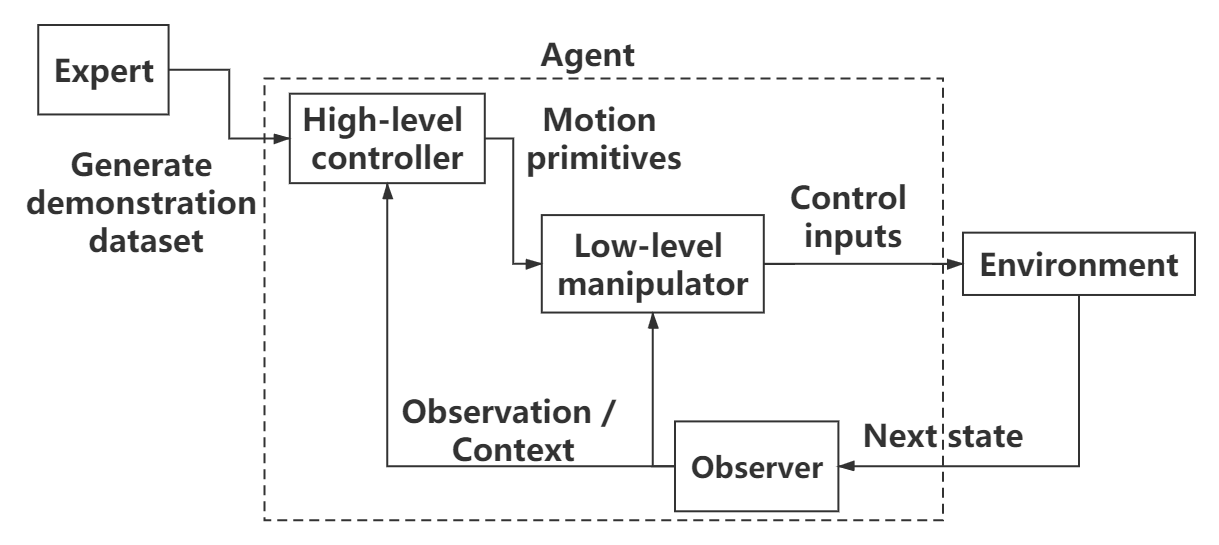}
      \caption{\textnormal{Control diagram adapted from \protect\cite{osaAlgorithmicPerspectiveImitation2018}}}
      \label{fig_h_l}
   \end{figure}

This subsection introduces a novel taxonomy, which divides IL into manipulation tasks and high-level tasks according to their evaluation approach. The idea is inspired by a control diagram (See Figure \ref{fig_h_l}) in\cite{osaAlgorithmicPerspectiveImitation2018}. Although some IL benchmark systems are proposed, such as\cite{lemmeOpensourceBenchmarkingLearned2015}, there is still no widely accepted one. In this case, the evaluation approaches and focus could vary from method to method, ranging from performance in sparse reward scenario to the smoothness of autonomous driving in dynamic environment.
This taxonomy could draw clearer boundary and might alleviate the difficulty of designing appropriate benchmark from performance perspective.
\begin{figure}[t]
      \centering
      \includegraphics[width=0.85\textwidth]{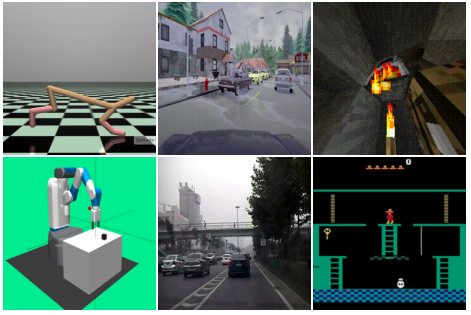}
      \caption{Prevalent Tasks in IL. Top-left: HalfCheetah in Mujoco; Top-mid: CARLA simulator; Top-right: Minecraft scenario in MineRL dataset; Bottom-left: FetchPickAndPlace-v1 in OpenAI Gym; Bottom-mid: Driving scenario in Xi'an \protect\cite{zhouModelingCarFollowingBehaviors2020}; Bottom-right: Atari game--MontezumaRevenge}
      \label{fig_task}
\end{figure}

The low-level manipulator tasks could be either real-world or virtual, and are not limited to robotics and autonomous driving problems. The robotic task can be object manipulation by robotic arm like PR2, KUKA robot arm, and simulation tasks commonly experimented on OPEN AI gym, MuJoCo simulation platform and so on. For real-world object manipulation tasks, the tasks could be push the object to the desired area, avoiding obstacles and operation soft object like rope. The autonomous driving tasks commonly implemented by simulation, and which is more related to the high-level planning. There are two widely-used benchmark system for simulation: CARLA CoRL2017 and NoCrash benchmark system, these two benchmark systems mainly focus on the urban scenario under various weather condition while the agent is evaluated on whether it can reach the destination on time, but CARLA CoRL2017 ignores the collision and traffic rules violation. Besides simulation, there are also some research doing experiment in real-world using cars\cite{zhouModelingCarFollowingBehaviors2020} and smaller remote-controlled cars\cite{codevillaEndtoEndDrivingConditional2018}, but other kinds of equipment are also used like remote control helicopter\cite{abbeelAutonomousHelicopterAerobatics2010}.
As for the high-level controller, the tasks could be navigation tasks and gameplay. The navigation tasks are mainly route recommendation and in-door room-to-room navigation. Most of the evaluated games are 2D Atari games on OpenAI Gym, such as MontezumaRevenge is commonly evaluated for performance on hard expolration and sparse reward scenario. Others are evaluated on 3D games like GTAV or Minecraft for evaluation. This taxonomy could be meaningful since it clearly reflects the target domain of the proposed algorithm, as the variance on their evaluation methods could be smaller, this may help to design a unified evaluation metric for IL. Figure \ref{fig_task} provides various popular evaluation tasks in IL. 

From the Figure \ref{fig_h_l}, the target of imitation could be either learning a policy for high-level controllers while assuming the low-level manipulator is working correctly or learning a policy to reproduce the simpler behavior on the low-level controller. Generally speaking, the high-level controller learns a policy to plan a sequence of motion primitives, such as \cite{osaOnlineTrajectoryPlanning2018}. As for the low-level controller, it learns a policy to reproduce the primitive behavior, such as \cite{sermanetTimeContrastiveNetworksSelfSupervised2018}, this forms the hierarchical structure of IL. Although some of the methods propose general frameworks which are evaluated on both domains, most of them are presenting ``bias" on selecting tasks to demonstrate their improvement in either higher-level or low-level domain. For example, in \cite{brownBetterthanDemonstratorImitationLearning2019}, the proposed algorithm is evaluated on both Atari and Mujoco environments, but the amount of the evaluated tasks in each environment is obviously unequal. In this case, the ambiguity of classifying these general methods could be simply eliminated based on their tendency on evaluation tasks.

\begin{table}[t]
\caption{Categorization of IL: Low-level Tasks vs. High-level Tasks}
\centering
\label{table_h_l}
\begin{tabular}{@{}cl@{}}
\toprule
Classes & Examples and Publications \\ \midrule
\multirow{4}{*}{Low-level manipulation}
&Surgical assistance\cite{osaOnlineTrajectoryPlanning2018,tanwani2020motion2vec}\\ \cmidrule(l){2-2} 
 & Vehicle manipulation\cite{zhouModelingCarFollowingBehaviors2020} \\ \cmidrule(l){2-2} 
 & Robotic arm\cite{sermanetTimeContrastiveNetworksSelfSupervised2018} \\ \cmidrule(l){2-2} 
 & VR teleoperation\cite{zhangDeepImitationLearning2018} \\ \midrule
\multirow{4}{*}{High-level tasks}&2D gameplay\cite{salimansLearningMontezumaRevenge2018} \\ \cmidrule(l){2-2} 
 & 3D gameplay\cite{arumugamDeepReinforcementLearning2019} \\ \cmidrule(l){2-2} 
 & Navigation\cite{husseinDeepImitationLearning2018} \\ \cmidrule(l){2-2} 
 & Sports analysis\cite{zhanGeneratingMultiAgentTrajectories2019} \\ \bottomrule
\end{tabular}
\end{table}

Table \ref{table_h_l} lists some of the recent research under this taxonomy. The majority of current imitation methods tend to use low-level manipulation tasks to evaluate the proposed method, since reinforcement learning performs acceptably in high-level controller tasks like games, and commonly performs poorly on the low-level manipulation tasks where the reward function might be impractical to obtain. Nevertheless, IL in the high-level controller tasks is non-trivial, since for the 3D tasks or hard exploration games, reinforcement learning can be time-consuming on the huge state and action space.

\subsection{BC vs. IRL vs. Adversarial Structured IL} 

This taxonomy is extended from the first taxonomy (BC vs. IRL). This new taxonomy divides IL into three categories: Behavioural Cloning (BC), Inverse Reinforcement Learning (IRL) and adversarial structured IL. 
With the recent development of IL, adversarial structured IL brings new insights for researchers and alleviate problems existing in previous work, such as high-dimensional problem. Inspired by the presence of GAIL, many recent papers adopt this adversarial structure, and inevitably, GAIL becomes baseline for comparison. But this is not enough to establish an independent category in IL, the true reason making it distinguishable is that GAIL is not belongs to either BC or IRL. Although adversarial structured IL has close connection with IRL, most adversarial structured IL does not recover the reward function. In this case, the taxonomy of IL could be more specific. GAIL and its derivations are separated from the traditional IRL category and classified as adversarial structured IL in this survey. Compared with the traditional taxonomies, the proposed new taxonomy is more adapted to the development of IL and eliminates the vagueness of classifying these adversarial structured methods.
\begin{figure}[t]
      \centering
      \includegraphics[width=0.9\textwidth]{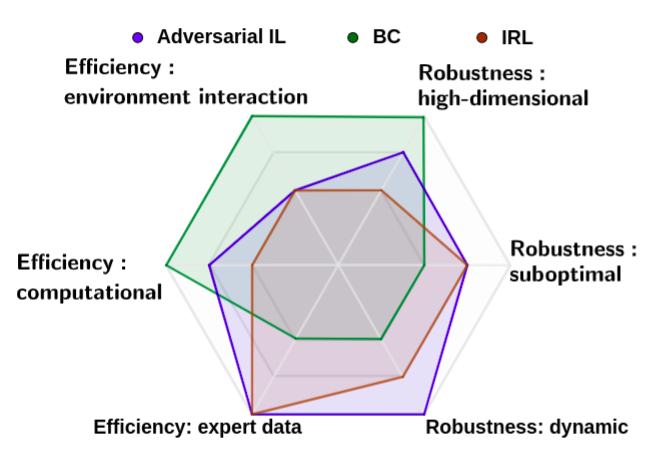}
      \caption{Web plot for taxonomy: BC vs. IRL vs. Adversarial Structured IL. We collected 6 popular evaluation criteria from the research and empirically ranked them into three levels based on research consensus. The outer the point, the higher the ranking, which means that it scores higher in the evaluation from the empirical perspective.}
      \label{fig_newtax}
\end{figure}

Figure \ref{fig_newtax} roughly evaluate the proposed three classes through two kinds of aspects which are commonly compared between research. Since different methods evaluate on various tasks, the overall performance is hard to quantify and rank, in this case, we evaluate three classes from Efficiency and Robustness from an empirical perspective. 

In terms of Efficiency, we mainly focus on environmental interaction, computation, and expert interaction. BC methods commonly take advantage of interaction with expert while have less interaction in the environment, and due to these characteristics, the computational cost for BC is more likely to be the lowest; IRL methods commonly have abundant interaction with the environment in their inner-loop, and the evaluation on system dynamic makes IRL suffers from high computational cost, but IRL methods hardly enquiry the expert during training; Adversarial structured IL methods also involve frequent interaction with the environment when they iteratively update the policy parameter and discriminator parameter, and get rid of the interaction with expert. As adversarial structured IL methods are commonly model-free, in the evaluation of computational efficiency, we rank it as the second.

In terms of Robustness, we mainly focus on robustness in high-dimensional space, robustness when demonstrations are suboptimal (includes the consideration on noise in demonstration), and robustness in dynamic system. BC methods commonly have better performance in high-dimensional space so that they are widely evaluation on robotics, while the performance in dynamic environment and suboptimal dataset are limited; IRL methods optimize the parameter in their inner-loop, which becomes a burden limiting their performance in high-dimensional space, but the recovered reward function would benefit the agent to do prediction in dynamic system. Since adversarial structured IL methods commonly derive from GAIL, they inherit the merits of GAIL: robustness in high-dimensional space and when changes occur in distribution. Because recent research such as\cite{dingGoalconditionedImitationLearning2019,brownExtrapolatingSuboptimalDemonstrations2019,Zuo2020offpolicyRAIL} in both IRL and Adversarial structured IL make progress in suboptimal demonstration problem, we give them the same rank in the evaluation of robustness on suboptimal demonstration.

\section{Main Research Topics and Methods}\label{main topic}
\subsection{Behavioural Cloning} 
Behavioural Cloning directly maps the states/contexts to actions/trajectories by leveraging the demonstration provided by expert/oracle. After generating the control input or trajectories, the loss function $\mathcal{L}$ will be designed according to the problem formulation and optimized in a supervised learning fashion. The state-of-the-art behavioural cloning uses negative log-likelihood loss to update the policy, i.e.
\[
\argmin_{\pi}\mathcal{L}(\pi)=-\frac{1}{N}\sum_{k=1}^{N}\log \pi(a_{k}|s_{k})
\]
Algorithm \ref{alg:BC} outlines the state-of-the-art behavioural cloning process. As traditional BC has less connection to MDP comparing with other prevalent methods, its efficiency is guaranteed, the trade-off is that it suffers from the scenario when the agent visits an unseen state.
Loss function $\mathcal{L}$ could be customized for specific problem formulation. Loss function (objective function) significantly influences the training process and there are many existing lost function available to measure the differences (in most cases, the difference means the 1 step deviation) such as $\ell_1$ loss, $\ell_2$ loss, KL divergence, Hinge Loss, etc.
For example, when using KL divergence as the loss function, the objective policy could be obtained by minimizing the deviation between expert distribution $q{\pi_{E}}$ and induced distribution $q(\pi)$, i.e.
\[
\pi^{*}=\argmin_{\pi}D_{KL}(q(\pi_{E})\|q(\pi)).
\]

BC could be subdivided into model-free BC and model-based BC methods. The main difference is whether the method learns a forward model to estimate the system dynamics. Since model-free BC methods take no consideration on the context, model-free BC methods perform well in industry applications where accurate controllers are available and experts could control and modify the robot joints. However, model-free BC methods typically are hard to predict future states and could not guarantee the output's feasibility under the environment that an accurate controller is not available. Under this kind of ``imperfect" environment, the agent would have limited information of system dynamics and usually gets stuck into the unseen scenarios due to the ``compounding error"\cite{rossEfficientReductionsImitation2010}. While model-based BC methods leverage the environment information and learn the dynamics iteratively to produce feasible output, the trade-off is that model-based BC methods usually have greater time-complexity since the iterative learning involvement process.

One of the significant BC method is DAgger, which is a model-free BC method proposed by Ross et al.\cite{rossReductionImitationLearning2011} and the idea is to use dataset aggregation to improve the generalization on unseen scenario. Algorithm \ref{alg:dagger} presents the abstract process of DAgger. DAgger adopts iterative learning process and mixes a new policy $\hat{\pi}^{n+1}$ with probability $\beta$ to construct the next policy. The mixing parameter is a set of $\{\beta_i\}$ that satisfies $\frac{1}{N}\sum_{i=1}^{N}\beta_{i} \rightarrow 0$.  
The start-up policy is learned by BC and records the trajectory into the dataset. Since a small difference can lead to compounding error, new unseen trajectories will be recorded combining with the expert's corrections. In this case, the algorithm gradually updates the possible state and fully leverages the presence of expert. Later research like\cite{rossReinforcementImitationLearning2014,sunDeeplyAggreVaTeDDifferentiable2017,liu2020imitationcache,hussein2021robust,wang2021minimizing,tu2021closing} were proposed to make improvements on DAgger.
This method alleviates the problem that traditional BC methods perform poorly on the unseen scenario and achieve data-efficiency comparing with previous methods like SMILe\cite{rossEfficientReductionsImitation2010}. However, it does have drawbacks, such as DAgger involves frequent interaction with the expert which might not be available and expensive in some cases (e.g., enquiring expert correction could be expensive in interdisciplinary tasks). Recent methods such as\cite{hoGenerativeAdversarialImitation2016,chenLearningCheating2019} successfully alleviate this problem. Another problem of DAgger could be that cost of each action is ignored. Since DAgger is evaluated on video games where the actions have equal cost, the cost of implementing each action is not obvious like tasks such as navigation tasks. This problem is solved later by Ross and Bagnell\cite{rossReinforcementImitationLearning2014}. 
\begin{algorithm}[t]
  \caption{Basic behavioural cloning method}
  \label{alg:BC}
  \begin{algorithmic}[1]
    \State Collect expert demonstration into dataset $\mathcal{D}$;
    \State Select policy representation $\pi_\theta$ and loss function $\mathcal{L}$;
    \State Use $\mathcal{D}$ to optimize the loss function $\mathcal{L}$ based on policy representation $\pi_\theta$;\\
    \Return optimized policy representation $\pi_\theta$;
  \end{algorithmic}
\end{algorithm}

\begin{algorithm}[t]
  \caption{DAgger \protect\cite{rossReductionImitationLearning2011}}
  \label{alg:dagger}
  \begin{algorithmic}[1]
    \State Initialize $\mathcal{D} \leftarrow \varnothing$; 
    \State Initialize $\hat{\pi}_{1}$ to any policy in $\Pi$;
    \For{$i = 1 \to N$}
        \State Let $\pi_{i} = \beta_{i}\pi^{*} + (1-\beta_{i})\hat{\pi}_{i}$.
        \State Sample T-step trajectory using $\pi_i$.
        \State \makecell[l]{Get dataset $\mathcal{D}_{i} = \{(s,\pi^{*}(s))\}$ of visited states by $\pi_i$\\ and action given by expert.}
        \State Aggregate dataset $\mathcal{D} \leftarrow \mathcal{D}\bigcup\mathcal{D}_i$.
        \State Train classifier $\hat{\pi}_{i+1}$ on $\mathcal{D}$.
    \EndFor\\
    \Return best $\hat{\pi}_{i}$ on validation.
  \end{algorithmic}
\end{algorithm}

\subsection{Inverse Reinforcement Learning}
Inverse reinforcement learning was firstly proposed by Russell\cite{russelllearning1998}. Unlike BC, the IRL agent is recovering and evaluating the reward function from expert demonstrations iteratively instead of establishing a mapping from states to actions. The choice of choosing BC or IRL depends on the problem settings. When the problem setting weights more on system dynamics and future prediction is necessary, choosing IRL methods can be more likely to evaluate the given context iteratively and provide a more accurate prediction. On the other hand, when an accurate controller and abundant demonstrations are available, choosing BC methods usually takes less time and performs better.   

IRL commonly assumes that the demonstrations are under Markov Decision Process setting and since the reward $\mathbb{R}$ is unknown, the set of states is used to estimate the feature vector (i.e. $\phi: \mathcal{X}\mapsto [0,1]^k$) instead of the true reward function (i.e. $\mathcal{X}\mapsto \mathbb{R}$). The process of classic IRL method (see Algorithm \ref{alg:IRL}) is based on iteratively update the reward function parameter $\omega$ and policy parameter $\theta$. The reward function parameter $\omega$ is updated after the state-action visitation frequency $u$ are evaluated, and the way that $\omega$ is updated could vary, for example, Ziebart et al.\cite{ziebart2008maximum} updated $\omega$ by maximizing the likelihood of the demonstration over maximum entropy distribution, i.e. $\omega^{*} = \argmax_{\omega}\sum_{\tau \in D} \log P(\tau\|\omega)$. On the other hand, the policy parameter $\theta$ is updated in the inner loop reinforcement learning process. This iterative and embedded structure can be problematic: the learning process could be time-consuming and impractical for high-dimensional problems like the high Degree Of Freedom (DOF) robotic problem. Another significant problem is ``ill-posed" which means the many different cost functions could lead to the same action. In this case, the good IRL methods need to have more expressive power and a more efficient framework. Research such as\cite{finnGuidedCostLearning2016,brownExtrapolatingSuboptimalDemonstrations2019,dasModelBasedInverseReinforcement2020,reddySQILImitationLearning2019,palanLearningRewardFunctions2019,ibarz2018reward,luo2021self} was proposed to alleviate the above problems by using more expressive models like neural network and optimizing the input like ranking the demonstration in advance.
\begin{algorithm}[t]
  \caption{Classic feature matching IRL method}
  \label{alg:IRL}
  \begin{algorithmic}[1]
    \Require The set of demonstrated trajectories $\mathcal{D}$;
    \State Initialize reward function parameter $\omega$ and policy parameter $\theta$;
    \Repeat
        \State \makecell[l]{Evaluate current policy $\pi_\theta$ state-action visitation \\frequency $u$;}
        \State \makecell[l]{Evaluate loss function $\mathcal{L}$ w.r.t. $u$ and the dataset $\mathcal{D}$\\ distribution;}
        \State \makecell[l]{Update the reward function parameter $\omega$ based on\\the loss function;}
        \State \makecell[l]{Update the policy parameter $\theta$ in the inner loop RL\\method 
                using the updated reward parameter $\omega$;}
    \Until\\
    \Return optimized policy representation $\pi_\theta$;
  \end{algorithmic}
\end{algorithm}

Several recent IRL methods are gradually integrated with various novel methods such as self-supervised learning. Self-supervised learning means learning a function from a partially given context to the remaining or surrounding context. Nair et al.\cite{nairCombiningSelfSupervisedLearning2017} could be one of the earliest researchers who adopt self-supervised learning into imitation learning. One important problem that integrating self-supervised learning with imitation learning has to solve is the huge amount of data, since the state and action space is extensive for real-world manipulation tasks. Nair et al. solved this problem by using the Baxter robot which automatically records data for a rope manipulation task. This method achieves practical improvement and provides a novel viewpoint for later research and leads the tendency of learning from the past. In 2018, Oh et al.\cite{ohSelfImitationLearning2018} proposed self-IL, which tries to leverage past good experience to get better exploration result. The proposed method takes a initial policy as input. It then iteratively uses the current policy to generate trajectories, calculates the accumulated return value $R$, update the dataset $D \leftarrow D \bigcup \{(s_t,a_t,R)\}_{t=0}^T$ and finally uses the deviation between accumulated return and the agent estimate value $R-V_\theta$ to optimize the policy parameter $\theta$. The process gradually ranks the state-action pairs and updates the policy parameter from the high-ranked pairs. In addition, Self-IL integrates Q learning with policy gradient under the actor-critic framework. As the component of the loss function, policy gradient loss was used to determine the good experience and lower bound Q learning was used to exploit the good experience, this helps Self-IL perform better in the hard exploration tasks. Similarly, in\cite{wang2019reinforced}, Self-supervised Imitation Learning (SIL) also tries to learn from its good experience but in a different structure. SIL creatively uses voice instruction in the imitation learning process. One language encoder is used to extract textual feature $\{\omega_i\}_{i=1}^n$ and an attention-based trajectory encoder LSTM is use to encode the previous state-action as a history context vector from visual state $\{v_j\}_{j=1}^m$, i.e. $h_t = LSTM([v_t,a_{t-1}],h_{t-1})$. Then visual context $c_t^{visual}$ and language context $c_t^{text}$ could be obtained based on the historical context vector, finally the action is predicted based on these parameters. The obtained experience is evaluated on a match critic, and the "good" experience is stored in a replay buffer for future prediction.

\subsection{Generative Adversarial Imitation Learning (GAIL)}
In order to mitigate problems in BC and IRL, Ho and Ermon\cite{hoGenerativeAdversarialImitation2016} proposed a novel general framework called Generative adversarial imitation learning in 2016. GAIL builds a connection between GAN\cite{goodfellow2014generative} and maximum entropy IRL\cite{ziebart2008maximum}. Inheriting from the structure of GAN, GAIL consists of a generative model G and a discriminator D, while G generates data distribution $\rho_\pi$ integrating with true data distribution $\rho_{\pi E}$ to confuse D. GAIL works in an iterative fashion, and the formal objective of GAIL could be denoted as 
\[
\min_{\pi}\max_{D\in(0,1)^{\mathcal{S}\times\mathcal{A}}}\hat{\mathbb{E}}_{\tau_i}[\log(D_{\omega}(s,a))]+\hat{\mathbb{E}}_{\tau_E}[\log(1-D_{\omega}(s,a))].
\]
GAIL firstly samples trajectories from initial policy, then these generated trajectories are used to update the discriminator weight $\omega$ by applying an Adam gradient step on equation
\[
\hat{\mathbb{E}}_{\tau_i}[\nabla_{\omega}\log(D_{\omega}(s,a))]+\hat{\mathbb{E}}_{\tau_E}[\nabla_{\omega}\log(1-D_{\omega}(s,a))],
\]
and maximize this equation with respect to D.
Then adopting the TRPO\cite{schulmanTrustRegionPolicy2017} with the cost function $\log(D_{\omega_{i+1}}(s,a))$ to update the policy parameter $\theta$ and minimize the above function with respect to $\pi$, combining with a causal entropy regularizer controlled by non-negative parameter $\lambda$, i.e. 
\[
\hat{\mathbb{E}}_{\tau_i}[\nabla_{\theta}\log\pi_{\theta}(a|s)\mathcal{Q}(s,a))]-\lambda\nabla_{\theta}H(\pi_\theta)
\]
\noindent 
where $\mathcal{Q}(\Bar{s},\Bar{a})=\hat{\mathbb{E}}_{\tau_i}[\log(D_{\omega_{i+1}}(s,a))|s_0 = \Bar{s},a_0 = \Bar{a}].$

\begin{algorithm}[t]
  \caption{GAIL \protect\cite{hoGenerativeAdversarialImitation2016}}
  \label{alg:gail}
  \begin{algorithmic}
  \Require Expert trajectories $\tau_E$ $\sim$ $\pi_E$, initial policy and discriminator parameter $\theta_0$, $\omega_0$
    \For{$i = 0,1,2,...$}
        \State Sample trajectories $\tau_i$ $\sim$ $\pi_{\theta_i}$.
        \State Update the discriminator parameters $\omega_i$ to $\omega_{i+1}$.
        \State Update the policy parameter $\theta_i$ to $\theta_{i+1}$.
    \EndFor
  \end{algorithmic}
\end{algorithm}
The abstract training process is presented in Algorithm \ref{alg:gail}. By adopting TRPO, the policy could be more resistant and stable to the noise in the policy gradient. Unlike DAgger and other previous algorithms, GAIL is more sample-efficiency from the perspective of using expert data and does not require expert interaction during the training process, it also presents adequate capacity dealing with the high-dimensional domain and changes in distribution. While the trade-off is the training process involves frequent interaction with the environment and could be more fragile and not stable for saddle point problem. As for the first problem, the authors suggested to initialize the policy with BC so that the amount of environment interaction would reduce. As for the second problem, recent research such as\cite{arenzNonAdversarialImitationLearning2020} tries to alleviate this problem by formulating the distribution-matching problem as an iterative lower-bound optimization problem.

\begin{table}[t]
\caption{Different Kinds of Derivative on GAIL}
\label{table_gail}
\centering
\begin{tabular}{@{}ll@{}}
\toprule
GAILs & Methods \\ \midrule
Make further improvement & MGAIL\cite{baramEndtoEndDifferentiableAdversarial2017}, InfoGAIL\cite{liInfoGAILInterpretableImitation2017} \\ \midrule
Apply to other research question & MAGAIL\cite{songMultiAgentGenerativeAdversarial2018}, GAIfO\cite{torabiGenerativeAdversarialImitation2019}\\ \midrule
Other generative model & Diverse GAIL\cite{wangRobustImitationDiverse2017}, GIRL\cite{yu2020intrinsic} \\ \bottomrule
\end{tabular}
\end{table}

Inspired by GAIL's presence, there is a bunch of research proposed to make further development on GAIL (see Table \ref{table_gail}) and adversarial structured IL gradually becomes a category.
In terms of ``make further improvement", many proposed methods modify and improve GAIL from different perspectives. For example, MGAIL\cite{baramEndtoEndDifferentiableAdversarial2017} uses an advanced forward model to make the model differentiable so that the Generator could use the exact gradient of the Discriminator. InfoGAIL\cite{liInfoGAILInterpretableImitation2017} modifies GAIL by adopting WGAN instead of GAN. Other recent work like GoalGAIL \cite{dingGoalconditionedImitationLearning2019}, TRGAIL\cite{kinoseIntegrationImitationLearning2020} and DGAIL\cite{zuoDeterministicGenerativeAdversarial2020} are all making improvement on GAIL by combining with other method like hindsight relabeling and Deep Deterministic Policy Gradient (DDPG) \cite{lillicrap2019continuous} to achieve faster convergence and better final performance.
In terms of ``apply to other research question", some of the proposed methods combine other method with GAIL and apply to various problems. For example, in\cite{sunProvablyEfficientImitation2019}, FAIL outperforms GAIL on sparse reward problem without using the ground truth action and achieves both sample and computational efficiency. It integrates adversarial structure with mini-max theory, which is used to determines the next time step policy $\pi_{h}$ under the assumption that $\left \{ \pi_{1}, \pi_{2},...,\pi_{h-1} \right \}$ is learned and fixed. GAIL is also applied into the other research area, such as multi-agent settings\cite{bhattacharyyaMultiAgentImitationLearning2018,songMultiAgentGenerativeAdversarial2018,zhanGeneratingMultiAgentTrajectories2019} and IfO settings\cite{torabiGenerativeAdversarialImitation2019} to effectively deal with more dynamic environment.
In terms of ``combine IL with other generative model", a number of recent research adopt other generative models to facilitate learning process, for example, in\cite{wangRobustImitationDiverse2017}, Variational AutoEncoder(VAE) is integrated with IL by using encoder to map from trajectories to an embedding vector $z$, which makes the proposed algorithm to behave diversely with relatively less demonstration and achieve one-shot learning for the new trajectory. Other research like GIRL\cite{yu2020intrinsic} also achieves the outstanding performance from limited demonstrations using VAE.

\begin{table}[t]
\caption{Publication Related to IfO}
\label{table_ifo}
\begin{center}
\begin{tabular}{@{}cl@{}}
\toprule
\textbf{Publication} & \textbf{Description}\\
\midrule
\makecell{IfO\cite{liuImitationObservationLearning2018}} & \makecell[l]{Learning policy from aligned observation only}\\
\midrule
\makecell{BCO\cite{torabi2018behavioral}} & \makecell[l]{Adopting IfO setting and integrating with BC}\\
\midrule
\makecell{TCN\cite{sermanetTimeContrastiveNetworksSelfSupervised2018}} & \makecell[l]{Multi-viewpoint self-supervised IfO method}\\
\midrule
\makecell{One-shot IfO\cite{aytarPlayingHardExploration2018}} & \makecell[l]{Extracting features from unlabeled and\\ unaligned gameplay footage}\\
\midrule
\makecell{Zero-Shot Visual Imitation\cite{pathakZeroShotVisualImitation2018}}  & \makecell[l]{Using distance between observations to predict and\\ penalize the actions}\\
\midrule
\makecell{IfO survey\cite{torabirecent2019}}& \makecell[l]{Detailed classified recent IfO methods}\\
\midrule
\makecell{Imitating Latent Policies\\ from Observation\cite{edwardsImitatingLatentPolicies2019}} & \makecell[l]{Infering latent policies directly from state observations}\\
\midrule
\makecell{GAIfO\cite{torabiGenerativeAdversarialImitation2019}} & \makecell[l]{Generative adversarial structure aggregating with IfO}\\
\midrule
\makecell{IfO Leveraging Proprioception\cite{torabiImitationLearningVideo2019}} & \makecell[l]{Leveraging internal information of the agent}\\
\midrule
\makecell{OPOLO\cite{zhu2021offpolicyifo}} & \makecell[l]{Using dual-form of the expectation function and\\ adversarial structure to achieve off-policy IfO}\\
\bottomrule
\end{tabular}
\end{center}
\end{table}

\subsection{Imitation from Observation (IfO)}
\label{ifo}
The prevalent methods introduced above is almost using sequences of state-action pairs to form trajectories as the input data. This kind of data preparation process could be laborious and this is a kind of waste for the abundant raw unlabeled videos. This problem got mitigated after IfO\cite{liuImitationObservationLearning2018} was proposed, and IL algorithms start to advocate this novel settings and make use of raw videos to learn policies. Comparing with traditional IL methods, this algorithm is more intuitive, and it follows the nature of how human and animal imitate. For example, people learn to dance by following a video, this kind of following process is achieved though detecting the changes of poses and taking actions to match the pose, which is similar to how IfO solves the problem. Different from traditional IL, the ground truth action sequence is not given. Similar to IRL, the main objective of IfO is the reward function from demonstration videos. Imitation from observation tries to build connection for different context so that the VAE structure is adopted to encode both the context (environment) of demonstrator (expert) $s_{1}$ and target context $s_{2}$. The proposed model has four components: a source observation encoder $Enc_1 (o_t^i)$ which extracts feature vector $z_1$, a target observation encoder $Enc_2 (o_0^j)$ which extracts feature vector $z_2$, a translator $z_3$ and a target context decoder $Dec(z_3)$. The model takes two sets of observations ($D_i = [o_t^i]_{t=0}^T$ and $D_j = [o_t^j]_{t=0}^T$ as source observation and target observation respectively) as input, then using these two sets to predict the future observation in target context under the assumption that source observation and target observation are time aligned. The translator $z_3$ translates features in $z_1$ produced by source encoder into the context of $z_2$ produced by another encoder, i.e. $z_3 = T(z_1, z_2)$, then the translated feature vector $z_3$ is decoded into the observation $\hat{o}_t^j$. The model is working in a supervised learning process with the loss function $\mathcal{L}_{trans} = \lVert (\hat{o}_t^j)-o_t^j\rVert_2^2$. To improve the performance, the final objective of the proposed model is combined with the loss of VAE reconstruction and the loss of time alignment, i.e.
\[
\mathcal{L} = \sum_{(i,j)}(\mathcal{L}_{trans}+\lambda_1 \mathcal{L}_{rec}+\lambda_2 \mathcal{L}_{align}),
\]
where $\lambda_1$ and $\lambda_2$ are the hyperparameter predetermined in advance.  The output reward function consists of two parts, the first one is deviation penalty on squared Euclidean distance, which measures the difference between the encoded learner's observation feature and translated expert observation feature in learner's context, i.e. 
\[
\hat{R}_{feat} (0_t^l) = -\lVert Enc_1 (o_t^l)-\frac{1}{n}\sum_{t=0}^T T(o_t^i,o_0^l)\rVert_2^2
\]The second part is the penalty which ensures the current observation keeping similar with translated observations, i.e. 
\[
\hat{R}_{img} (0_t^l) = -\lVert o_t^l-\frac{1}{n}\sum_{t=0}^T M(o_t^i,o_0^l)\rVert_2^2,
\] 
where M is the full observation translation model. The proposed reward function could be applied into the any reinforcement learning algorithm, Liu et al. uses TRPO\cite{schulmanTrustRegionPolicy2017} for the simulation experiments. 

After IfO being proposed, measuring observation distance to replace the ground truth action becomes a prevalent setting in imitation learning. In Table \ref{table_ifo}, we present some of the research advocate this new insight and apply this idea into various domain. Both BC, IRL and GAIL start to adopt this setting to simplify the input. For example, in\cite{aytarPlayingHardExploration2018}, raw unaligned YouTube videos are used for imitation to reproduce the behavior for games. YouTube videos are relatively noisy and varying in settings like resolution. The proposed method successfully handled these problems by using a novel self-supervised objective to learn a domain-invariant representation from videos. Similarly, in\cite{sermanetTimeContrastiveNetworksSelfSupervised2018}, multi-viewpoint self-supervised IL method Time-Contrastive Network (TCN) was proposed. Different viewpoints introduce a wide range of contexts about the task environment and the goal is to learn invariant representation about the task. By measuring the distance between the input video frames and ``looking at itself in the mirror", the robot could learn its internal joint to learn the mapping and achieve imitating demonstration. 

\section{Challenges and Opportunities}\label{future}
Although improvements like integrating novel techniques, reducing human interaction during training and simplifying inputs alleviate difficulties in learning behaviour, there are still some open challenges for IL:

\textbf{Diverse behavior learning: } \textnormal{Current IL methods commonly use task-specific training datasets to learn to reproduce single behavior. Research like\cite{wangRobustImitationDiverse2017} presented diverse behavior learning by combining adversarial structure and variational autoencoder, but this is still an open challenge. Other methods could be adopted to optimize IL, such as transfer learning might help the agent to learn from similar tasks so that the training process could be more efficient.}

\textbf{Sub-optimal demonstration for training: } \textnormal{Current IL methods generally require a high-quality set of demonstrations for training. However, the number of high-quality demonstrations could be limited and expensive to obtain. Existing research like\cite{brownExtrapolatingSuboptimalDemonstrations2019,dingGoalconditionedImitationLearning2019,songweakly2020} have shown the possibility to use sub-optimal demonstration for training, but performance can be improved by extracting common intent from the dataset. }

\textbf{Imitation not just from observation: } \textnormal{Current IfO methods commonly use raw videos and the deviation of observations to recover the reward function. But the video is not just observation, maybe the voice instruction could also be used to get a better reward function. Wang et al.\cite{wang2019reinforced} demonstrated using natural language for navigation tasks, but it could be an interesting topic to explore in the IfO settings.}

\textbf{Better representation:} \textnormal{Good policy representation could benefit the training process to achieve data-efficiency and computation-efficiency. Finding better policy representation is still an active research topic for IL. Besides policy representation, how to represent the demonstration is another problem in IL. The representation of demonstration needs to be more efficient and expressive. }

\textbf{Find globally optimal solution:} \textnormal{Most research is finding a locally optimal solution based on demonstration, which might set the upper-bound for the agent performance. The future direction could be finding the global optimal for a specific task, which requires the agent to understand the intent of the behavior instead of copy-pasting. Current research like\cite{yu2020intrinsic} successfully surpasses the demonstrator's performance, but finding the global optimal still needs effort.}

\section{Conclusion}
Imitation learning achieves outstanding performance in a wide range of problems, ranging from solving hard exploration Atari games to achieving object manipulation while avoiding obstacles by robotic arm. Different kinds of imitation learning methods make contribution to this significant development, such as BC methods replicate behavior more intuitively where the environmental parameters could be easily obtained; IRL methods achieve data-efficiency and future behavior prediction when problems weight more on environment dynamics and care less about training time; adversarial structured IL methods eliminate expert interaction during the training process and present adequate capacity dealing with the high-dimensional problem. While IL methods continue to grow and develop, IL is also seeking breakthroughs in settings, like IfO methods simplify the input by replacing the need of action labels when the input demonstrations are raw video. Although recent work presents a superior advantage in replicating behavior, taxonomy ambiguity exists as the presence of GAIL and its derivatives break out of the previous classification framework. To alleviate this ambiguity, we analyzed the traditional taxonomies of IL and proposed new taxonomies that draw clearer boundaries between methods. Despite the success of  IL, challenges and opportunities exist, such as diverse behavior learning, leveraging sub-optimal demonstration and voice instruction, better representation, and finally finding the globally optimal solution. Future work is expected to unravel IL and its practical applications.

\bibliographystyle{ACM-Reference-Format}
\bibliography{sample-base}

\end{document}